\documentclass[conference]{IEEEtran}
\usepackage{graphics}
\usepackage{graphicx}

\usepackage{url}
\usepackage{amsmath}
\usepackage{float}
\usepackage{refstyle}
\usepackage{caption}

\begin{document}
%
\title{Discovering conversational topics and emotions associated with Demonetization tweets in India}

\author{\IEEEauthorblockN{Mitodru Niyogi}
\IEEEauthorblockA{
Govt. College of Engineering \& Ceramic Technology \\
Kolkata, West Bengal 700010\\
Email: mitodru.niyogi@gmail.com}
\and
\IEEEauthorblockN{Asim Kumar Pal}
\IEEEauthorblockA{Management Information Systems\\
Indian Institute of Management Calcutta, Kolkata\\
Email: asim@iimcal.ac.in}
}


%


\maketitle

\begin{abstract}
	
	Social media platforms contain great wealth of information which provides
	us opportunities explore hidden patterns or unknown correlations, and understand people's satisfaction with what they are discussing. As one showcase,
	in this paper, we summarize the data set of Twitter messages related to recent demonetization of all Rs. 500 and Rs. 1000 notes in India and explore insights
	from Twitter's data. Our proposed system automatically extracts the popular latent topics
	in conversations regarding demonetization discussed in Twitter via the Latent
	Dirichlet Allocation (LDA) based topic model and also identifies the correlated topics
	across different categories. Additionally, it also discovers people's opinions expressed
	through their tweets related to the event under consideration via the emotion
	analyzer. The system also employs an intuitive and informative visualization to
	show the uncovered insight. Furthermore, we use an evaluation measure, Normalized
	Mutual Information (NMI), to select the best LDA models. The obtained LDA
	results show that the tool can be effectively used to extract discussion topics and
	summarize them for further manual analysis.
	
\end{abstract}


%
\IEEEpeerreviewmaketitle

\section{Introduction}\label{sec:Introduction}
Analyzing news stories have been pivotal for finding out some of the quantitative and qualitative attributes from text documents. A broad domain like news analytics incorporates the use of various text mining methods to analyze text. It applies methods from Natural Language Processing, Machine Learning, Information Retrieval, etc. In our study the qualitative attributes can be socio-economic tags related to demonetization in India. The sentiment score which generally reflects the tone (positive/negative) of the text as well as the emotions expressed, can be one of the quantitative attributes. 
In this paper we have dealt with two problems in the domain of news analytics; firstly is text categorization \cite{sri} without any prior domain knowledge i.e., topic modeling and secondly is emotion analysis. For example we are trying to investigate how emotions of people relate to demonetization in India.

For text categorization, we have clustered the news stories into several k topics: unsupervised learning with automatic topic labeling i.e., topic modeling \cite{topm}. Topic modeling reflects the thematic structure of the collection of documents by treating data as observations which gets derived from a generative probabilistic process that comprises hidden variables for documents. Inferring them using posterior inference results the topics generation that describes its corpus.  The emotion analysis (also referred as sentiment extraction) \cite{sa} would give an emotion association score to each story depending on the expressive tone of the story in 8 basic emotions categories and two sentiments (positive/negative) deciding the tone of the overall story.
Introduction is here.
The roadmap of the paper is as follows. Data preparation and exploratory insights are described in section 2. Section 3 is on background. Section 4 reveals our proposed system architecture. Section 5 deals with the experiment setup. Section 6 gives the results. Section 7 draws conclusions from discussions and points to future work.
\section{Data }
\label{sec:2}

\subsection{Data Set}
The data has been collected over a period of two months from November 13 to December 18, 2016 across four metro cities: Delhi, Kolkata, Mumbai, and Chennai based on sets of keywords corresponding to demonetization in India (e.g., ``demonet'', ``black money'', ``cashless'', etc.) using Twitter's streaming API \cite{api} and was stored into mongoDB \cite{mongo}. We approximately collected 73,970 tweets \cite{tweet} in the order of retweet count during the period. Novel data comprise extraction date and time, user ID, user name, tweets message, and geographical area. Due to the huge volume of novel data, we divide the data into only with dates, user IDs, and text and conduct further operation and analysis based on the three variables. Most of the tweets are written in English, but the original raw data set also includes the tweets in vernacular languages such as Hindi or Bengali. We did exclude them in the initial data manipulation process. The data from the nosql database was imported into R console using the tm package \cite{tm} in CRAN library to construct the document term matrix for use in developing the topic model.

\subsection{Exploratory Insight}
We explore the time series analysis of tweets over given time period. We visualize the number of retweet by hour, minute; average number of words by hour. We also explore which users have contributed to maximum tweets in our corpus. It also determines user's influence over others in terms of his retweet count. Out of 73,970 tweets we see that most of the tweets are from Twitter Web Client source followed by Windows phone, iPhone sources. We see that more than 10 users have tweeted more than 100 tweets for the event under consideration.

 Fig. 1 shows the hourly retweet count, Fig. 2 displays hourly average count of words in tweets, Fig. 3 shows the top 7 source contributors (platforms) for generation of tweets, Fig. 4 lists the top Twitter handlers with maximum tweets count, Fig. 5 and Fig. 6 show some word clouds of the corpus.
\begin{figure}[H]
	\centering
	\includegraphics[width=\linewidth]{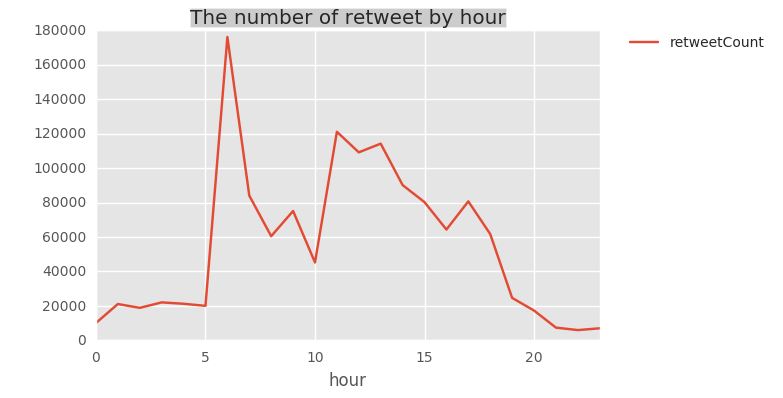}
	\caption{Hourly retweet}
\end{figure}

\begin{figure}[H]
	\centering
	\includegraphics[width=\linewidth]{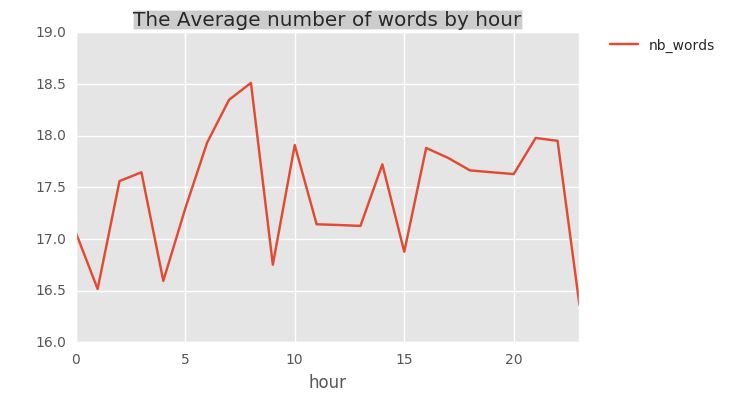}
	\caption{Average number of words hourly}
\end{figure}

\begin{figure}[H]
	\centering
	\includegraphics[width=\linewidth]{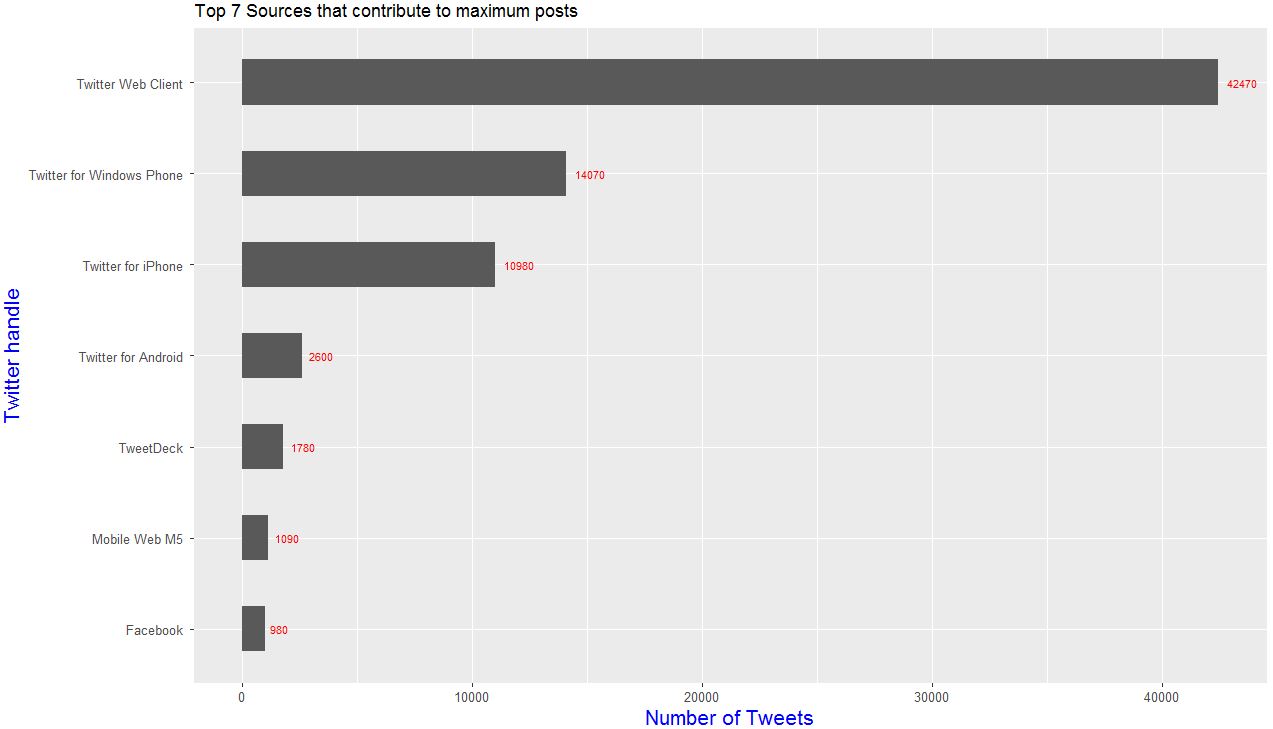}
	\caption{Top 7 sources contribution to maximum tweets}
\end{figure}

\begin{figure}[H]
	\centering
	\includegraphics[width=\linewidth]{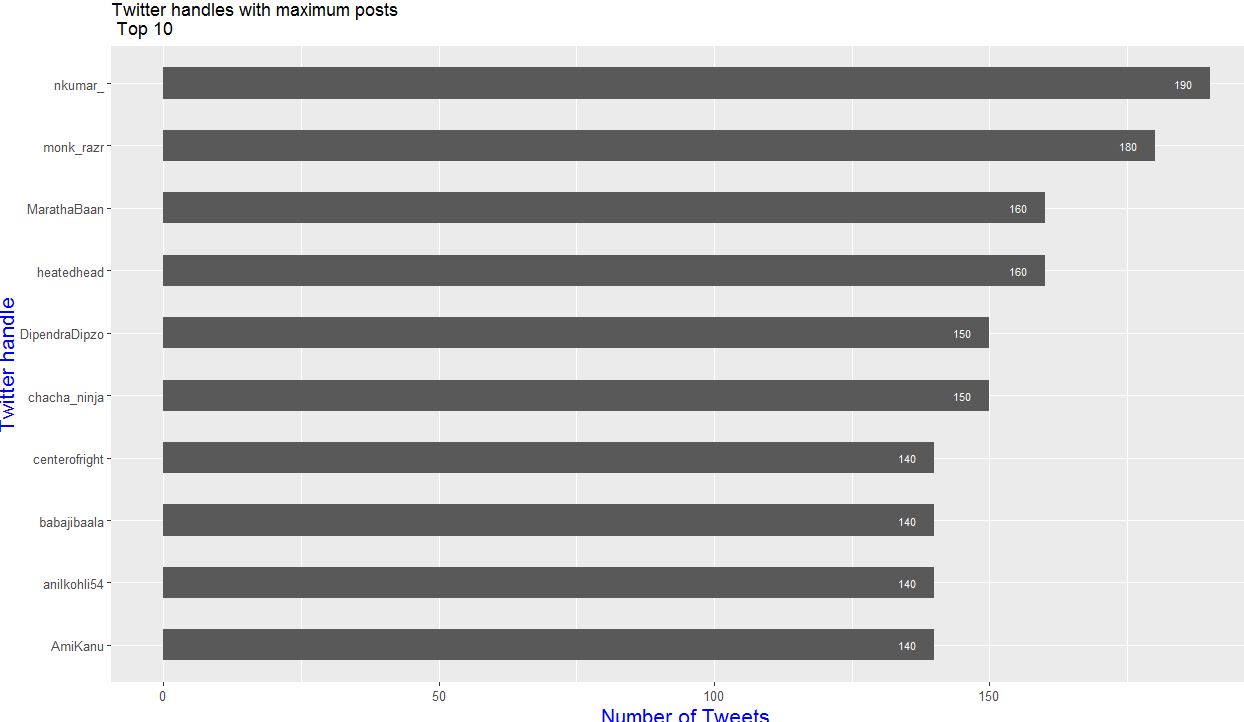}
	\caption{Top 10 Twitter handles with maximum tweets}
\end{figure}

\begin{figure}[H]
	\centering
	\includegraphics[width=\linewidth]{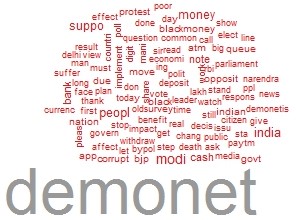}
	\caption{Wordcloud of corpus having high frequency words}
\end{figure}

\begin{figure}[H]
	\centering
	\includegraphics[width=\linewidth]{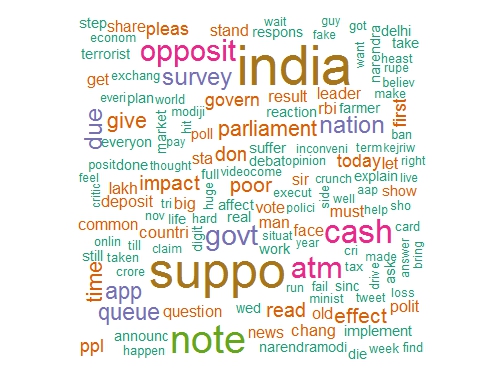}
	\caption{Wordcloud of corpus after removing set of keywords used to download data}
\end{figure}

\subsection{Data Preprocessing}
Before applying any of the sentiment/emotion extraction methods, we perform data preprocessing. Data preprocessing allows to produce higher quality of text classification and reduce the computational complexity. Typical preprocessing procedure includes the following steps:

\textbf{Stemming and lemmatization}. Stemming is a procedure of replacing words with their stems, or roots. The dimensionality of the Bag-Of-Words representation is reduced when root-related words, such as ``read", ``reader" and ``reading" are mapped into one word ``read".
Over stemming lowers precision and under-stemming lowers recall. The overall impact of stemming depends on the dataset and stemming algorithm. The most popular stemming
algorithm is Porter stemmer \cite{port}.

\textbf{Stop-words removal}. Stop words are words which carry a connecting function in the sentence, such as prepositions, articles, etc. \cite{sal} There is no definite list of stop words, but some search machines, are using some of the most common, short function words, such as ``the", ``is", ``at", ``which" and ``on". These words can be removed from the text before classification since they have a high frequency of occurrence in the text, but do not affect the final sentiment of the sentence.

\textbf{TF-IDF model}. Term Frequency Inverse Document Frequency (TF-IDF) \cite{yat} divides the term frequencies by the document frequencies (number of documents were the j$^{th}$ word has appeared). This adjustment is done in order to lower the weightage of those words which are common across all the documents. The TF-IDF measure suggests how important the term is for the particular document. In TF-IDF scheme words which are common across all documents will automatically get less importance.\\

\let\labelitemi\labelitemii

Preprocessing of tweet include following points,
\begin{itemize}
	\item Remove all URLs (e.g. www.xyz.com), hash tags (e.g. \#topic), targets ($@$username)
	
	\item Correct the spellings; sequence of repeated characters is to be handled
	
	\item Replace all the emoticons with their sentiment.
	
	\item Remove all punctuations ,symbols, numbers
	
	\item Remove Stop Words
	
	\item Remove Non-English Tweets
	
\end{itemize}



\section{Background}
\label{sec :3}

\subsection{Introduction to LDA}
Previously, documents were treated as ``a-bag-of-words'' \cite{zou} approach as in many models which dealt with text documents. Topic modeling adopts that a document is ``a-bag-of-topics'' instead of ``a-bag-of-words'' representation, and its sole purpose is to cluster each term in each document into a relevant topic. A variations of different probabilistic topic models \cite{em} have been proposed and LDA \cite{lda} is considered to be a well known method. Alike other methods, the input to LDA is a term $\times$ document matrix, and the output of LDA is composed of two distributions, namely document-topic distribution $\theta$ and topic-word distribution $\phi$.
EM \cite{topic} and Gibbs Sampling \cite{gibbs} algorithms were proposed to derive the distributions of $\theta$ and $\phi$ . In this paper, we use the Gibbs Sampling based LDA. In this approach, one of the most significant step is updating each topic assignments individually for each term in every documents according to the probabilities calculated using Equation 1. 
\begin{equation}
P( \, z_{i}=k\ \mid w_{i}=v,z_{-i},w_{-i}) \, \propto \frac{C_{vk}^{WT} + \beta}{\Sigma _{v ^{'}} \, C ^{WT} _{vk} + V \beta} \cdot \frac{C_{dk}^{DT} + \alpha}{\Sigma _{k ^{'}} \, C ^{DT} _{dk} + K \alpha}\;
\end{equation}

where z$_{i}$=k represents that the i$^{th}$ term in a document is assigned to topic k, w$_{i}$ =v is the mapping of the observed term w$_{i}$ to the v$_{th}$ term in the corpus's vocabulary, and z$_{-i}$ signifies all the assignments of topic except the i$_{th}$ term. C$^{WT}_{vk}$ is the frequency of occurrence of term v assigned to a particular topic k, and C$^{DT}_{dk}$ is the number of times that the document d contains the topic k. Moreover, K is the user input denoting the number of topics, V represents the vocabulary’s size, hyper-parameters for the document-topic distribution and topic-word distribution are denoted by $\alpha$ and $\beta$ respectively. By default, $\alpha$ and $\beta$ are set to 50/K and 0.01.

We perform N iterations of Gibbs sampling for every terms in the corpus and after this, we estimate the document-topic $\theta$ and topic-word $\phi$ distributions respectively using Equations 2 and 3.

\begin{equation}
\theta _{dk}=\frac{C_{dk}^{DT} + \alpha}{\Sigma _{k ^{'}} \, C ^{DT} _{dk} + K \alpha}
\end{equation}

\begin{equation}
\phi _{vk}= \frac{C_{vk}^{WT} + \beta}{\Sigma _{v ^{'}} \, C ^{WT} _{vk} + V \beta}
\end{equation}

%
%
%
%


%
%
\subsection{Emotion Analysis}
\begin{figure}[H]
	\centering
	\includegraphics[width=\linewidth]{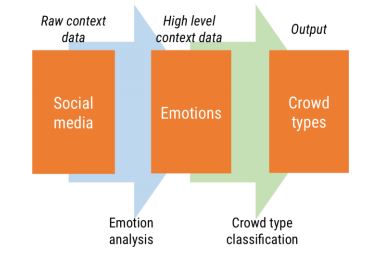}
	\caption{The process to identify a crowd type from social media via emotion analysis}
	\label{fig 1: emotion}
\end{figure}

Emotion classification is fundamentally a text classification problem. Traditional sentiment classification mainly classifies documents as positive, negative and neutral. In this scenario, the emotion is fine-grained into basic emotions such as anger, fear, anticipation, trust, surprise, sadness, joy, and disgust. In this paper, the NRC Word-Emotion Association Lexicon Corpus \cite{nrc} is selected as the labeled corpus. It comprises a list of English words and their associations with Plutchik's \cite{emo} eight basic emotions and two sentiments (negative and positive). It involves three variables `TargetWord', `AffectCategory', and `AssociationFlag'. TargetWord is a word for which emotion associations are provided. AffectCategory is one of eight emotions or one of two polarities (negative or positive). AssociationFlag has one of two possible values: 0 or 1.  0 indicates that the target word has no association with affect category, whereas 1 indicates an association. Fig. \ref{fig 1: emotion} shows the process to identify a crowd type from social media.

\subsection{NMI}
In the experiment we used NMI (Normalized Mutual Information) \cite{pool} to evaluate overall documents (tweets) cluster quality. The following formula is used to calculate NMI:
\begin{equation}
NMI(X,Y)=\frac{2I(X,Y)}{H(X)+H(Y)}
\end{equation}
where I(X;Y) is mutual information between X and Y, where X = {X1, X2, ...Xn} and Y = {Y1, Y2,...Yn}. Xi is the set of text reviews in LDA's topic i while Yj is the set of text reviews with the label j. In our experiments, a text review with the label j means that the text review has the highest probability of belonging to topic j; n is the number of topics. I(X;Y) is 
\begin{equation}
I(X,Y)=\sum_{y\in Y} \sum_{x \in X} p(x,y) \log(\, \frac{p(x,y)}{p(x)p(y)}\,)\,\;
\end{equation}
In the formula, p(x$_{i}$) means probability of being classified to topic i, p(y$_{j}$) means
probability of labeled to topic j while p(x$_{i}$,y$_{j}$) means probability of being classified to cluster i but actually labeled to cluster j. H(X) is entropy of X as calculated by the
following formula:

\begin{equation}
H(X)=-\sum_{i=1}^{n}p(x_{i})\log p(x_{i})
\end{equation}
The clustering result is totally different from the label if the value of NMI is 0 and is identical if value of NMI is 1.
\section{Proposed System Architecture}

\begin{figure}[H]
	\centering
	\includegraphics[width=\linewidth]{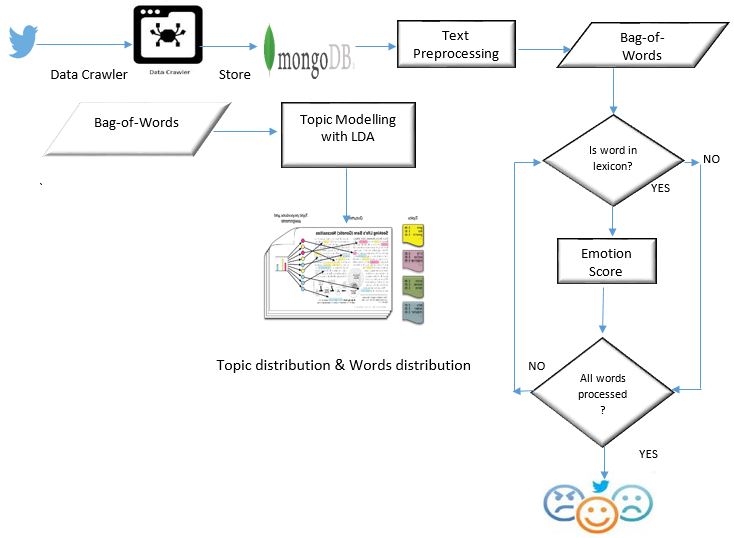}
	\caption{Proposed System Architecture}
	\label{system}
\end{figure}

We propose a system that consists of three main components including data collection, data analysis and data visualization.  The data collection module is developed to crawl the tweets from Twitter using data crawlers and to store the tweets into MongoDb, a NoSQL database for scalability and scheme less data storage purpose. After data preprocessing steps such as tokenization, stemming and stopwords removal, the system mainly performs two different types of analyses to answer the following questions:
\begin{itemize}
	
	\item What are the topics discussed by people online to help us understand people's interests?
	\item What are people's opinion on the specific topics to help us understand their satisfaction of those topics?
\end{itemize}
The term-document matrix is created which is fed to LDA based model for discovering latent topics and the documents are analyzed by the emotion analyzer. Then, emotion analyzer will tag each tweet as happy, sad, angry, fear, surprise or neutral. Fig. \ref{system} presents the architecture of our proposed system. \ 

\section{Experimental Setup}
\label{sec:4}
For the Demonetization data, we started with default parameters $\alpha$ = 0.1; $\beta$ = 0.01 and input parameter topic number N =5, 10, 15, 20 which means 5, 10, 15, 20 desired topics. By
comparing the LDA result given in Table \ref{tab2}, we choose topic number N = 15 as a basic group for further comparison since when N = 15, most topics have enough words to reveal information about the topic while without too much words to make the topics messy. In the next step of our experiment, we set N = 15 and tuning parameter $\alpha$ and $\beta$ by setting $\alpha$ = 0.1, 0.05, 0.2 while $\beta$ = 0.01, 0.015, 0.007 to see if the results show any difference. 

We performed the Emotion Analysis using syuzhet \cite{syu} CRAN package which is based on NRC Emotion Lexicon on the dataset. As a result, 73,970 tweets were labeled with one of eight emotions: anger, anticipation, disgust, fear, joy, sadness, surprise, trust and two sentiments (positive and negative) to determine the overall tone of the event.



\section{Results} %
\label{sec:5}

\subsection{Discovering Topics}
\begin{figure}[H]
	\begin{center}$
		\begin{array}{ccccc}
		\includegraphics[width=0.2\linewidth]{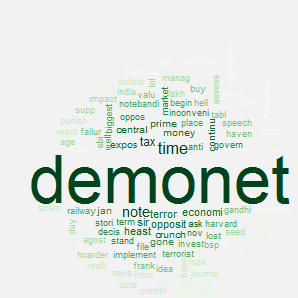} 
		\includegraphics[width=0.2\linewidth]{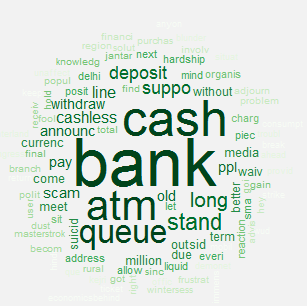}
		\includegraphics[width=0.2\linewidth]{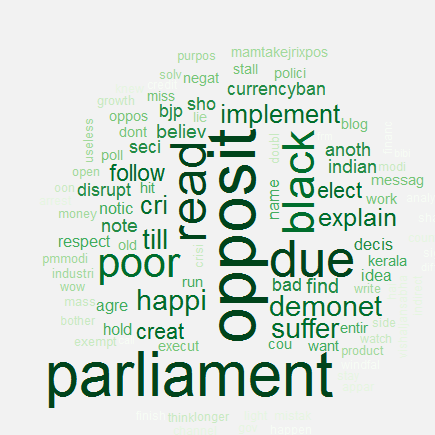}
		\includegraphics[width=0.2\linewidth]{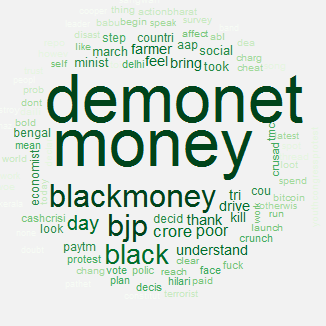}
		\includegraphics[width=0.2\linewidth]{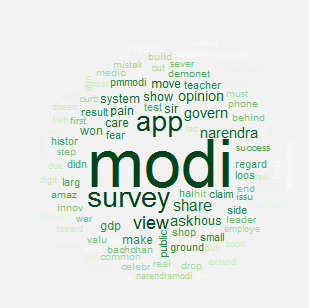}
		\end{array}$
	\end{center}
	\caption{Topics}
	\label{Topics}
\end{figure}

Table \ref{tab:1} shows a list of 5 topics. The words are arranged in the order of highest probability of words distribution to topics. Fig. \ref{Topics} displays some topics word cloud. 

\begin{figure}[H]
	\centering
	\includegraphics[width=\linewidth]{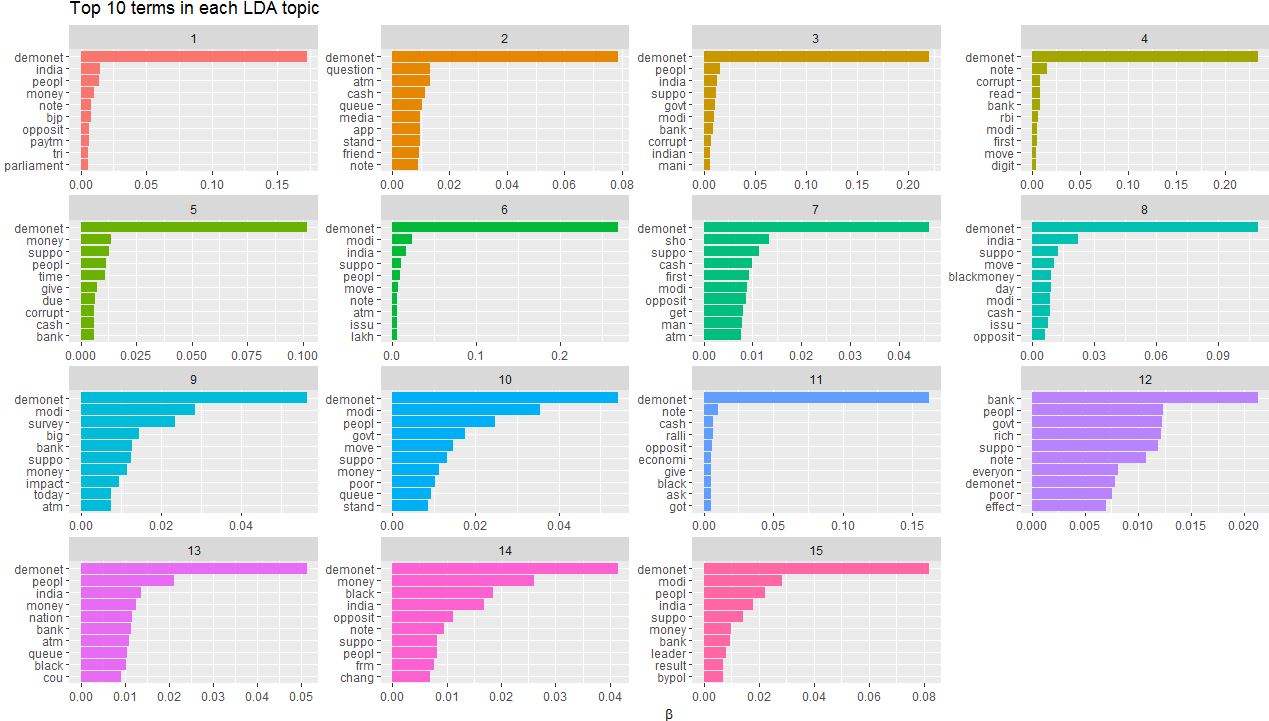}
	\caption{Top 10 terms in 15 topics}
	\label{Top terms}
\end{figure}

Topic 1 lists ``bank'', ``queue'', ``atm'', ``stand''. This reflects the hectic issues related to bank/ATM transaction. Topic` the impact of currency ban on life of citizens which has led to deaths. Topic 4 reveals parliamentary debate on demonetization. Topic 5 reflects farmer and opposition parties protest. Topic 6 indicates people's support for demonetization. Topic 7 lists words ``don'', ``modi'', ``rbi'', ``impact'', looks like a mixed topic. Topic 8 lists ``modi'', ``fights'', ``corrupt'', ``leader'', ``blackmoney''. This indicates people's support and acknowledgment of PM Modi's decision. Topic 9 lists ``Kashmir'', ``protest''. Topic 10 discusses about impact on terror funding due to note ban. Topic 11 portrays currency ban as a vote bank politics supported by the govt as it lists ``bypol'', ``farmer'', ``congress'', ``affect'', ``move'', ``bjp'' words. Topic 12 indicates huge economic and job loss. Topic 13 tells about harassment of people due to this event as aggressive words such as ``disgust'', ``harass'' dominate. Topic 14 talks about cash crunch in banks as it lists ``cashless'', ``rbi'', ``crunch''. Topic 15 tells about encouraging online transactions as it lists ``app'',``paytm'', ``easy'', ``online''. Fig. \ref{Top terms} shows the distribution of top 10 terms in collection of 15 topics.

\begin{table}
	\caption{Discovering Topics: 5 topics from 15 topics}
	\label{tab:1}       
	%
	%
	\begin{tabular}{p{1.25cm}p{1.25cm}p{1.25cm}p{1.25cm}p{1.25cm}}
		\hline\noalign{\smallskip}\centering
		Topic 1 & Topic 2 & Topic 3 & Topic 4 & Topic 5  \\
		\noalign{\smallskip}\hline\noalign{\smallskip}
		bank & demonet  & nation & paytm  & opposit \\
		atm & stop & currenc & demonet  & protest\\
		cash & blackmoney& suffer & parliament & govt \\
		queue & benefit & impact & debate & farmer \\
		stand & citizen & life & corrupt & affect\\
		long & plan & death & modi & poor\\
		\noalign{\smallskip}\hline\noalign{\smallskip}
	\end{tabular}
	
\end{table}

\subsection{NMI Results}

In our experiments, we evaluated NMI of LDA with different topic numbers. Table \ref{tab2} reports the results:



\begin{table}[H]
	\centering
	\caption{NMI of models}
	\label{tab2}
	
	\begin{tabular}{|l|l|}
		\hline
		\multicolumn{1}{|c|}{LDA Models}          & NMI Results \\ \hline
		Demone(5,0.1,0.01)                        & 0.548       \\ \hline
		Demone(10,0.1,0.01)                       & 0.588       \\ \hline
		\multicolumn{1}{|c|}{Demone(15,0.1,0.01)} & 0.590       \\ \hline
		Demone(20,0.1,0.01)                       & 0.463       \\ \hline
	\end{tabular}
\end{table}

The results show that with fewer topics, the NMI value tends to be higher. Since NMI presents similarity of clustered tweets set and labelled tweets set, the overall NMI results indicate that with fewer topics, tweets set are more correctly clustered. The
reason for this phenomenon could be the length of each document (\,tweet\,) is much shorter if compared to traditional documents. Since the length for each tweet is limited (\,usually no longer than 140 characters\,), information contained in a single tweet is also limited. Hence, when the number of topics increases, many topics tend to contain the same words; as a result, it is hard to determine to which topic a document be assigned.
In further experiments, we can use different tweeter pooling schemes \cite{pool} and see whether they affect the NMI results.

\subsection{Emotion Count}
\begin{figure}[H]
	\centering
	\includegraphics[width=\linewidth]{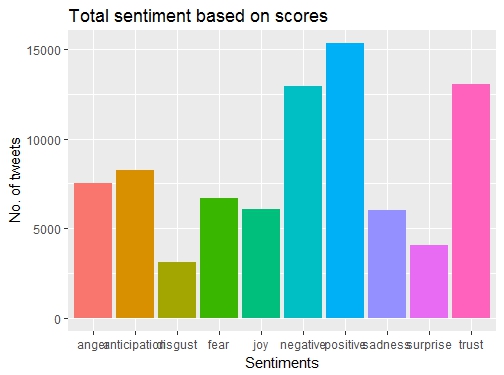}
	\caption{Emotion distribution during the event}
	\label{fig:4}
\end{figure}

Fig. \ref{fig:4} shows the distribution of emotions during this event. As can be seen, the dominating emotion is trust followed by anticipation and anger. The reason might be that due to the mixed reactions of people expressing their thoughts and opinions through tweets. More than 12,500 tweets express trust as an emotion. Around 8000 tweets express anticipation. 7000 tweets express fear, with a count of around 7500 tweets of anger emotion, around 3000 tweets are of disgust and 6000 tweets express sadness. Disgust emotion was the least emotion expressed in our study. More than 15000 tweets express positive sentiment and around 13000 indicate negative sentiments.

\section{Discussion and Conclusion}
\label{sec:6}

As substantial number of people are connected to online social networking services, and a noteworthy amount of information related to experiences, and practices in consumption is shared in this new media form. Text mining is an emergent technique for mining valuable information from the web especially related to social media. Our objective is to discovering tweets semantic patterns in users' discussions and trend on social media about demonetization in India.

In order to detect conversations in connection to the event under consideration, we applied Latent Dirichlet Allocation based probabilistic system to discover latent topics. We varied the LDA parameters to find a model whose output is more informative as evaluated by NMI. Performance of the LDA models were not affected by changes in distribution parameters $\alpha$ and $\beta$. At the same time, the results significantly changed with the change of topic numbers. As we expected, the quality of LDA results also depends on the amount of records in the data. Manual analysis of the results revealed that LDA is able to extract most of the detailed information from the data. It extracts all the major event components, including the people involved, how the event unfolded etc. However, in some topics we can't infer to a specific label due to its mixed nature. It is also important to note that all the extracted topics are related to the events covered by the collected data. Our method not only confides to the analysis of case study presented but also significant to the analysis of Twitter data collected in similar settings.  From our analysis, we observed that the positive response has exceeded the negative aspects about the demonetization discussion as shown in the emotion distribution plot in Fig.~\ref{fig:4} which also does not rule out large section of people have raised voices against the event. Trust, anticipation and anger are the top 3 emotions in count which reflects that our study is not biased towards one polarity.

Understanding the influence of social networks can help government agencies to better understand how such information can be used not only in the dissemination of a socio-economical event, but can also help to draw responses that could help to mitigating an unruly reaction or preventing violence from starting and escalating.

\end{document}